\documentclass[twoside,11pt]{article}
\usepackage[preprint]{dmlr2e}
\usepackage{amsmath}
\usepackage{tabularx}
\usepackage{array}
\usepackage{booktabs}
\usepackage{url}
\makeatletter\g@addto@macro\UrlBreaks{\do\_\do\a\do\b\do\c\do\d\do\e\do\f\do\g\do\h\do\i\do\j\do\k\do\l\do\m\do\n\do\o\do\p\do\q\do\r\do\s\do\t\do\u\do\v\do\w\do\x\do\y\do\z\do\A\do\B\do\C\do\D\do\E\do\F\do\G\do\H\do\I\do\J\do\K\do\L\do\M\do\N\do\O\do\P\do\Q\do\R\do\S\do\T\do\U\do\V\do\W\do\X\do\Y\do\Z}\makeatother
\DeclareUnicodeCharacter{00D7}{\ensuremath{\times}}
\DeclareUnicodeCharacter{00B7}{\ensuremath{\cdot}}
\DeclareUnicodeCharacter{00B2}{\textsuperscript{2}}
\DeclareUnicodeCharacter{2264}{\ensuremath{\leq}}
\DeclareUnicodeCharacter{2265}{\ensuremath{\geq}}
\DeclareUnicodeCharacter{2212}{\ensuremath{-}}
\setcitestyle{numbers,square,comma,sort&compress}
\providecommand{\tightlist}{\setlength{\itemsep}{0pt}\setlength{\parskip}{0pt}}
\usepackage{lastpage}
\dmlrheading{}{2026}{1-\pageref{LastPage}}{11/26}{}{}{Achraf Toufah, Rana Saad and Moulay Youssef El Hafidi}
\ShortHeadings{A Pre-training Acceptance Gate for Mammography Datasets}{Toufah, Saad and El Hafidi}
\firstpageno{1}
\makeatletter
\def\@maketitle{\vbox{\hsize\textwidth
 \linewidth\hsize \vskip \beforetitskip
 {\begin{center} \Large\bf \@title \par \end{center}} \vskip \aftertitskip
 {\def\and{\unskip\enspace{\rm and}\enspace}%
  \def\addr{\small\it}%
  \def\email{\hfill\small\sc}%
  \def\name{\normalsize\bf}%
  \def\AND{\@endauthor\rm\hss \vskip \interauthorskip \@startauthor}
  \@startauthor \@author\@endauthor}
  \vskip \aftermaketitskip
}}
\makeatother
\def\openreview{}

\begin{document}

\title{Label leakage through file headers: geometry and batch confounds in public mammography datasets, and a pre-training acceptance gate}

\author{\name Achraf Toufah \email ORCID 0009-0006-6802-3905\\
       \addr Qanary Inc.
       \AND
       \name Rana Saad \email ORCID 0009-0005-5806-2604\\
       \addr Qanary Inc.
       \AND
       \name Moulay Youssef El Hafidi \email myelhafidi@qanary.net\\
       \addr Qanary Inc. (corresponding author; ORCID 0000-0001-7413-3293)}

\editor{To be assigned}

\maketitle

\begin{abstract}%
A dataset whose diagnostic label can be recovered from non-diagnostic information yields high, reproducible and uninformative model performance. We applied pre-training controls that see only image geometry, global texture, a sub-collection identifier or one covariate to six public mammography datasets, against a stop threshold of AUC 0.60. In KAU-BCMD, geometry alone predicted the BI-RADS class at AUC 1.0000 in a local copy and at a mean of 0.979 in a public Kaggle archive; after resizing, texture alone gave 0.847. In CMMD, the sub-collection prefix predicted malignancy at 0.7753; the restricted sub-collection passed geometry (0.5000) and texture (0.5810) but failed an identifier-order control added post hoc (test AUC 0.9926). CBIS-DDSM exceeded the stop on geometry (pooled mean 0.654 over 20 splits), a verdict statistic chosen post hoc; its official mass and calcification splits did not. M-SYNTH, INbreast and a partial VinDr-Mammo copy passed the few controls run. We release the code and derived tables. There is no quantum component in the inference path.
\end{abstract}


\section{Introduction}

A machine-learning result on medical images is a statement about anatomy only if the label cannot be obtained in some other way. When the label is also recoverable from something that carries no diagnostic content, such as the width of the file, the sharpness left by an export step, or the position of a record in a list, every model trained on that dataset can score well, the scores replicate, and the replication says nothing about disease. The general phenomenon has been described as shortcut learning in deep neural networks \citep{r1} and as ``Clever Hans'' prediction \citep{r2}, and dataset bias was examined in computer vision earlier still \citep{r3}.

Medical imaging has supplied concrete cases. A deep-learning model for detecting pneumonia in chest radiographs showed variable generalisation performance in a cross-sectional study \citep{r4}. In pelvic radiographs, a deep-learning model predicted hip fracture using confounding patient and healthcare variables: the scanner model could itself be predicted from the image at an AUC of 1.00, and fracture prediction fell to chance on a test set in which fracture risk was balanced across patient and hospital-process variables \citep{r5}. The association between surgical skin markings in dermoscopic images and the diagnostic performance of a convolutional network for melanoma recognition has been studied \citep{r6}. Hidden stratification has been reported to cause clinically meaningful failures in machine learning for medical imaging \citep{r7}. For radiographic COVID-19 detection, models were reported to select shortcuts over signal \citep{r8}, and the methods proposed for automatic COVID-19 detection from X-ray images have been evaluated critically \citep{r9}. Methodological failures of this kind in machine learning for medical imaging have been surveyed together with recommendations \citep{r10}.

Two neighbouring literatures describe the same defect in other terms. Leakage has been linked to a reproducibility crisis in machine-learning-based science \citep{r11}. In high-throughput biology, batch effects arise because measurements are affected by laboratory conditions, reagent lots and personnel, and they become a major problem when the batch is correlated with the outcome of interest \citep{r12}; bias across sites has likewise been detected and corrected in multi-site neuroimaging datasets \citep{r13}. In mammography specifically, Samala et al.~simulated feature leakage in the classification of breast masses and showed that validation performance could be strongly overestimated relative to an independent test set \citep{r14}; Elmejgari et al.~quantified patient-level leakage on CBIS-DDSM, including a protocol in which the official training and test splits are concatenated before cross-validation \citep{r15}; Liao and Aagaard published an open codebase aimed at transparency in deep-learning diagnosis on CBIS-DDSM \citep{r16}; Mabrok compared side-level with patient-level evaluation on CMMD \citep{r17}; Zufiria et al.~analysed potential biases in mammography datasets for deep-learning model development \citep{r18}; and Garrucho et al.~studied domain generalisation of mass detection in a multi-centre study \citep{r19}.

Reporting practice has responded mainly through documentation and checklists: datasheets for datasets \citep{r20}, the FAIR guiding principles for scientific data management and stewardship \citep{r21}, and a checklist for artificial intelligence in medical imaging intended as a guide for authors and reviewers \citep{r22}. Large public medical image datasets have been explored after their release \citep{r23}, and a review of eight machine-learning datasets in mammography against the FAIR principles found that they vary considerably, mixing digital captures with scanned film and differing in licensing, ease of access, labelling reliability and file formats \citep{r24}. Documentation describes a dataset; it does not by itself test whether the label can be predicted from the file.

The datasets examined here are six: the King Abdulaziz University Breast Cancer Mammogram Dataset (KAU-BCMD) \citep{r25}; the Chinese Mammography Database (CMMD) \citep{r26,r27}, distributed by The Cancer Imaging Archive \citep{r28}; CBIS-DDSM, the curated subset of the Digital Database for Screening Mammography \citep{r29}; INbreast \citep{r30}; VinDr-Mammo \citep{r31}; and the simulated M-SYNTH cohort \citep{r32}. Such datasets are also pooled: one benchmark compilation collates images from six sources, among them DDSM, INbreast, KAU-BCMD and CMMD \citep{r33}. Pooling is the situation in which a batch that is correlated with the label matters most.

We report what a small set of file-level controls returned on these six datasets, examined in several configurations (KAU-BCMD as a local copy and as a public Kaggle archive; CMMD pooled and restricted to one sub-collection). Three datasets fail at least one stop-role control. In KAU-BCMD, where the BI-RADS category is encoded as directory membership, a classifier given only image width, height, megapixels and aspect ratio separates BI-RADS 4/5 from BI-RADS 1 at AUC 1.0000 in a local copy and at a mean of 0.979 over 20 patient-grouped splits in a public Kaggle archive; no image dimension is shared between the classes in either copy, and after resizing to a common 224 × 224 the texture control still gives 0.847 on the public archive. In CMMD, the sub-collection prefix of the patient identifier alone predicts malignancy at AUC 0.7753. In CBIS-DDSM, geometry alone is associated with pathology at a mean AUC of 0.654 in the pooled cohort, 0.688 in the calcification cohort and 0.635 in the mass cohort; that verdict rests on a mean over repeated splits which was chosen after the results were seen, and on the single official splits the mass and calcification cohorts are below the stop. Three datasets pass every control that was run on them, and few were run: M-SYNTH (two controls), INbreast (geometry only, 0.486) and the available 59.8\% subset of VinDr-Mammo (geometry only, 0.573).

CMMD restricted to its first sub-collection is in neither group. It passes the registered geometry and texture controls and then fails a control we added afterwards: the serial number in the patient identifier alone predicts malignancy at a test AUC of 0.9926. We report this as the paper's own demonstration that a pass on the registered controls is not a certificate.

Relative to the work cited above, this paper adds measurements, a procedure and a release, and no new method. The measurements are control AUCs for named public datasets and copies, including the unremarkable ones. The procedure is to fix a small set of impoverished controls and a threshold in writing before a model of interest is trained, and to record which analytic choices were made after results were seen. The release is the code, the notebooks with their outputs and the derived tables. We make no priority claim. We queried the CrossRef bibliographic index with the seven search strings listed in Methods and inspected the first six records returned for each; these included a report of patient-level leakage on CBIS-DDSM \citep{r15} and an evaluation study on CMMD \citep{r17}, and none of them, judged by its title, reported a geometry--label association in KAU-BCMD or CBIS-DDSM or an association between the CMMD identifier and malignancy. This was not a systematic review, and records beyond the first six per query were not examined. Two preprints, identified by a web search on the same day and read at abstract level, are closely related. Hajishafiezahramini et al.~\citep{r34} found that a classifier separated the dataset of origin almost perfectly across three mammography sources that included CBIS-DDSM and CMMD, after identical preprocessing, and that adding externally sourced positive cases lowered screening performance. Zafari et al.~\citep{r35} describe a harmonisation pipeline and quantify distributional differences across CBIS-DDSM, a CMMD derivative and VinDr-Mammo. Both address differences between datasets; the controls reported here ask whether the label is recoverable from non-diagnostic information within one dataset.

The work arose while benchmarks comparing quantum with classical machine-learning models were being set up on clinical breast images, and that context explains why the controls were run at all. Had they not been run on the local KAU-BCMD copy, we would expect every arm of the planned benchmark, quantum and classical alike, to have scored near AUC 1.0, and a quantum arm to have been reported at roughly 0.99; such a result would have been reproducible, internally consistent and uninformative. A recently proposed framework for quantum advantage \citep{r36}, which is a proposal and not a standard, makes validated correctness of the output the first of its two criteria; a dataset whose label is readable from its file header fails that requirement for every arm equally, before any computation begins. This paper reports no quantum advantage, no speed-up and no quantum-enhanced performance. Every quantum run in the wider programme to date is a classical statevector simulation. There is no quantum component in the inference path.

\begin{center}\rule{0.5\linewidth}{0.5pt}\end{center}

\section{Methods}

\subsection{Controls and stop rule}

Each control is a classifier trained on an input that should carry no diagnostic information, scored by the area under the receiver operating characteristic curve (AUC) \citep{r37} against the diagnostic label. A control is expected to score about 0.50.

\begin{table}[!tbp]
\centering\footnotesize
\caption{\textbf{The controls.} ``Stop'' controls halt a study when they exceed the threshold; the covariate control is reported as context only.}
\label{tab:1}
\begin{tabularx}{\textwidth}{>{\raggedright\arraybackslash}X>{\raggedright\arraybackslash}X>{\raggedright\arraybackslash}X>{\raggedright\arraybackslash}X>{\raggedright\arraybackslash}X}
\toprule
\textbf{Control} & \textbf{Input} & \textbf{Role} & \textbf{Threshold} & \textbf{Status} \\
\midrule
Geometry & image width, height, megapixels, aspect ratio; no pixel content & stop & about 0.50 expected; 0.60 stops the study & registered before the CMMD run \\
Texture & Laplacian variance, gradient energy, intensity standard deviation, contrast range, after a common resize; no geometry & stop & about 0.50 expected; 0.60 stops the study & registered before the CMMD run \\
Batch / site & the acquisition-batch or sub-collection identifier alone & stop & about 0.50 expected; 0.60 stops the study & added after the CMMD texture control failed \\
Single covariate & one demographic variable alone, for example age & context only, never a stop & none & registered before the CMMD run \\
Identifier order & the numeric (serial) part of the patient identifier alone & stop & about 0.50 expected; 0.60 stops the study & added post hoc, after the registered controls had passed CMMD D1 \\
\bottomrule
\end{tabularx}
\end{table}

The classifier for every control is scikit-learn's \nolinkurl{HistGradientBoostingClassifier} with default hyperparameters and \nolinkurl{random_state = 0}. It is deliberately untuned: a tuned control would invite the objection that the gate had been optimised to fail.

The stop threshold of 0.60 was fixed in writing before any control was run on CMMD. The registered sentence states that the geometry and texture controls must return approximately 0.50 AUC on the frozen test split and that the benchmark stops and the dataset is investigated if either ``exceeds 0.60''. The follow-up notebooks implement the rule as a fail when the mean AUC is greater than or equal to 0.60. The two formulations differ only at the boundary. No value reported in this paper lies at the boundary, so no verdict depends on the difference.

The threshold is a screening rule, not a statistical test. It attaches no p-value or error rate to a verdict, and it is not offered as a universal boundary between acceptable signal and problematic leakage: a control below it does not show that a dataset is free of shortcuts, and a control above it does not measure how far a downstream model would exploit them. Its role is operational, deciding whether training proceeds or the dataset is investigated first. Three properties support that use. First, the controls carry no diagnostic content, so their expected AUC is 0.50, and 0.60 lies well outside the split-to-split variation of the controls that returned values near chance in this study: the largest standard deviation over 20 repeated splits was 0.0217 (INbreast geometry, the smallest dataset), which places 0.60 at 4.6 standard deviations above 0.50, and no single split of the four controls with means within 0.05 of chance exceeded 0.5570. Second, under a binormal model with equal variances an AUC of 0.60 corresponds to a standardised mean difference of 0.36, a small effect, so the rule flags label information that is modest in size but larger than the sampling variation seen here. Third, every verdict was re-scored at thresholds of 0.55, 0.65, 0.70 and 0.75, and the conclusions that do and do not depend on the choice are stated in Section 5.

Two features of the design matter. The geometry control decodes no pixels; it reads four integers from each file header. The covariate control is reported but never used as a stop, because an association between a demographic variable and the outcome can be a property of the clinical population and not a defect of the dataset.

\subsection{Registration status and deviations}

The registration referred to throughout is an internal, dated written protocol. It was not deposited in a public registry. The protocol and its dated deviations log are part of the release (Data availability).

The geometry, texture and single-covariate controls and the 0.60 threshold were registered for CMMD before any model was trained. The KAU-BCMD controls on the local copy were not registered: they were run as an ad hoc diagnostic during benchmark set-up, and the gate was written into the protocol afterwards as a consequence of what they found. The KAU-BCMD result on the local copy is therefore exploratory. The batch/site control was promoted to a stop role after the CMMD confound had been found by investigation. The identifier-order control was added later still (Section 2.6). The M-SYNTH controls were run under the protocol of a separate study.

Two further items were registered before they were run, and were carried out in a follow-up round of analyses with the following deviations. Throughout, ``initial'' refers to the first round of controls (Table 2) and ``follow-up'' to the second round, scored over repeated splits (Table 3).

\begin{enumerate}
\def\labelenumi{\arabic{enumi}.}
\tightlist
\item
  A fresh download of the public KAU-BCMD distribution was registered for two publisher routes under an institutional account. It was run on one route only, a Kaggle-hosted archive downloaded by a team member and held in a cloud copy, and it was scored under 20 repeated patient-grouped splits instead of one.
\item
  Gates on five further public datasets were registered: CBIS-DDSM, VinDr-Mammo, INbreast, the RSNA screening-mammography challenge dataset \citep{r38} and BreastMNIST \citep{r39}, each with four controls and a bit-depth check on a frozen split. Three of the five were run (CBIS-DDSM, INbreast, VinDr-Mammo), with the geometry control only, and each was scored under 20 repeated patient-grouped five-fold splits; where an official split exists, its single-split AUC is reported beside the mean.
\end{enumerate}

The 0.60 threshold and the 20-split design were fixed before the follow-up runs. The decision to apply the threshold to the mean over the 20 splits, and not to a single split, was taken after the results were seen. It matters for CBIS-DDSM (Section 3.5).

\subsection{Datasets and copies analysed}

\begin{table}[!tbp]
\centering\footnotesize
\caption{\textbf{Datasets and copies analysed in the initial round.} }
\label{tab:2}
\begin{tabularx}{\textwidth}{l>{\raggedright\arraybackslash}X>{\raggedright\arraybackslash}X>{\raggedright\arraybackslash}X}
\toprule
\textbf{\textasciitilde{}} & \textbf{KAU-BCMD \citep{r25}, labelled PNG folder (local copy)} & \textbf{CMMD \citep{r26,r27}} & \textbf{M-SYNTH \citep{r32}} \\
\midrule
Label & BI-RADS category, encoded as directory membership; no pathology ground truth in the copy & benign/malignant by biopsy, in a separate metadata file & lesion present/absent by construction \\
n & 1,991 PNG images · 501 cases · 1,000 case-side units & 1,775 patients · 1,872 breasts · 5,202 images & 1,200 cases \\
Class balance & 60 BI-RADS 4--5 breasts against 938 BI-RADS 1 & 1,316 malignant / 556 benign breasts (70.3\% malignant) & balanced by construction \\
Format & 8-bit PNG, non-uniform image sizes & MONOCHROME2, 2294 × 1914 (4.39 MP); 5,200 images at 8 bits and 2 at 16 & simulated, 256 × 256 patches \\
Licence & CC BY 4.0 & CC BY 4.0 & CC0 \\
Scoring & one patient-grouped five-fold cross-validation (\nolinkurl{StratifiedGroupKFold}) & once on a frozen test split & once on a frozen test split \\
\bottomrule
\end{tabularx}
\end{table}

The dataset paper for KAU-BCMD states that the dataset was annotated and reviewed by three radiologists \citep{r25}. The dataset paper for CMMD describes two branches, one with biopsy-confirmed benign or malignant labels and one with known molecular subtypes \citep{r26}; the two-part structure of CMMD is therefore documented by its authors and is not a finding of this work. That paper reports fewer mammograms than the number of image files in the archive manifest we analysed. We did not reconcile the two counts, and every CMMD count in this paper is taken from the manifest and the label file.

\begin{table}[!tbp]
\centering\footnotesize
\caption{\textbf{Datasets and copies analysed with the geometry control in the follow-up round.} }
\label{tab:3}
\begin{tabularx}{\textwidth}{l>{\raggedright\arraybackslash}X>{\raggedright\arraybackslash}X>{\raggedright\arraybackslash}X>{\raggedright\arraybackslash}X}
\toprule
\textbf{\textasciitilde{}} & \textbf{KAU-BCMD \citep{r25}, public Kaggle archive} & \textbf{CBIS-DDSM \citep{r29}} & \textbf{INbreast \citep{r30}} & \textbf{VinDr-Mammo \citep{r31}} \\
\midrule
Source & Kaggle-hosted archive \nolinkurl{king-abdulaziz-university-mammogram-dataset.zip} (606,558,154 bytes, members dated 2022-03-01) & The Cancer Imaging Archive \citep{r28}, official download client & original distribution, ``INbreast Release 1.0'' & PhysioNet; download incomplete \\
Label & BI-RADS folder; 1 against 4/5 & pathology from the official case-description files; malignant against benign (including benign without callback) & BI-RADS 1/2 against 4/5 (assessment, not pathology) & breast-level BI-RADS 1/2 against 4/5 (assessment, not pathology) \\
n analysed & 1,991 JPEG images · 501 patients (BI-RADS 1/4/5) & 2,853 full-field images · 1,460 patients & 379 images · 103 patients & 11,404 images · 2,993 studies \\
Format & 8-bit JPEG & 16-bit DICOM (digitised film) & 16-bit DICOM & 12- and 16-bit DICOM \\
Licence & CC BY 4.0 & CC BY 3.0 plus the archive's data-usage policy & access agreement with the providers; the agreement is not on file with the authors & PhysioNet Restricted Health Data License 1.5.0 (credentialed access) \\
Coverage & complete (2,378 of 2,378 images in the archive) & 37 of 284 official calcification-test breast-views present; all other official sets at least 99.7\% & complete (410 of 410 images) & 11,965 of 20,000 images (59.8\%) \\
\bottomrule
\end{tabularx}
\end{table}

Patients were grouped as follows: KAU-BCMD by the year-and-case identifier in the file name, with a sensitivity analysis grouping by case number only; CBIS-DDSM by the patient identifier across both cohorts; INbreast by the anonymised patient identifier in the file name; VinDr-Mammo by study identifier as a proxy for the patient.

The scoring protocols are not identical across datasets and we do not present them as equivalent. The local KAU-BCMD copy was cross-validated because no frozen split was ever created for it: the dataset was stopped before splitting. CMMD and M-SYNTH were scored once on a frozen split. The controls available also differ, because the recorded metadata differ: M-SYNTH has no patient-identifier prefix and no age, so batch and covariate controls do not exist for it, and its texture control is a 15-feature whole-image physics bank and not the four-feature bank used on the clinical datasets.

\subsection{Preprocessing remedies for KAU-BCMD}

Three standard remedies were applied in turn and the control re-scored: resizing every image to a common 224 × 224; equalising to a common 467-pixel short axis and then resizing; and restricting to a resolution-matched subset below 0.5 megapixels. In the two resizing conditions the geometry features are constant by construction, so the control used is texture only.

The initial analysis named the four texture statistics without defining them numerically. For the follow-up re-run on the public archive they were fixed as follows: 8-bit greyscale; bilinear resampling; Laplacian variance as the variance of the 4-connected discrete Laplacian; gradient energy as the mean of dx² + dy² by central differences; intensity standard deviation as the pixel standard deviation; contrast range as the 99th minus the 1st percentile. The local and public texture values are therefore not demonstrably computed with identical definitions.

\subsection{Frozen splits for CMMD}

The CMMD splits are stratified by the patient's worst finding and grouped by patient, with seed 20261020, and were written to disk before any model was trained. The unit of analysis is the breast, because 30 of the 97 patients with two labelled breasts have a different classification on each side. After the texture control failed on the pooled dataset the protocol was amended to the first sub-collection (D1) and the splits were re-frozen, before any model was trained. The released split files are these D1 splits: 605 training, 171 validation and 331 test breasts, the test split holding 168 malignant and 163 benign breasts.

\subsection{Identifier-order control (post hoc)}

CMMD patient identifiers have the form \nolinkurl{D1-0001}. After the registered gate had passed CMMD D1, the numeric part of the identifier was used as the only feature of the same untuned classifier. It was scored on the frozen D1 validation and test splits, and over 20 repeated patient-grouped five-fold splits of the 1,107 D1 breasts. This control was not one of the four registered controls. It is exploratory, and it is labelled post hoc wherever it appears.

The same control was subsequently applied to CBIS-DDSM, whose patient identifiers have the form \nolinkurl{P_00016}. Labels were taken from the four public case-description files of the archive \citep{r28}, and no image was read. The unit was the image (patient, side and view within a cohort), positive if any lesion on it was malignant, as in the geometry control. It was scored over 20 repeated patient-grouped five-fold splits for each cohort and for both cohorts pooled, grouping by patient across cohorts, and on the official training and test files of each cohort. It was not run on INbreast or VinDr-Mammo, whose copies are held in cloud storage that was not accessible for this addition, or on KAU-BCMD, which already fails on geometry.

\subsection{Statistics}

The initial controls are each reported as a single AUC, run once at \nolinkurl{random_state = 0}. The follow-up controls are reported as the mean, standard deviation (SD), minimum and maximum of the out-of-fold AUC over 20 repeated patient-grouped five-fold splits (\nolinkurl{StratifiedGroupKFold} with shuffling and split seeds 0--19; the model is fixed at \nolinkurl{random_state = 0} because the classifier is deterministic for a given training set), with the verdict taken on the mean. No p-value is reported for any control and no multiplicity correction is applied: the gate is a screen against an absolute threshold and not a test of a null hypothesis, and correcting for multiplicity would make a safety screen more permissive. No dispersion is available for the initial controls.

The follow-up analyses ran under Python 3.13.5 with scikit-learn 1.9.1, pandas 3.1.0, numpy 2.5.4, pydicom 3.0.2 and Pillow 13.0.0. The software versions of the initial analyses were not recorded.

\subsection{No quantum component}

No circuit, simulated or otherwise, is constructed or evaluated anywhere in this study. The gate is a gradient-boosting classifier on impoverished inputs.

\subsection{Literature search}

The statement in the Introduction about earlier reports rests on queries to the CrossRef bibliographic index made during the preparation of this article, with the first six records inspected for each of these strings: ``KAU-BCMD mammogram dataset classification deep learning''; ``CMMD Chinese Mammography Database data leakage batch sub-collection''; ``CBIS-DDSM data leakage patient overlap training test split''; ``shortcut learning mammography spurious correlation image size dataset bias''; ``duplicate images public mammography dataset INbreast CBIS-DDSM leakage''; ``label leakage file metadata DICOM header medical imaging dataset''; and ``image resolution confound deep learning medical image dataset class label image size''.

\begin{center}\rule{0.5\linewidth}{0.5pt}\end{center}

\section{Results}

\subsection{Overview}

Table 4 lists every control result. Figure 2 shows the same rows.

\begin{figure}[!tbp]\centering
\includegraphics[width=0.8\textwidth]{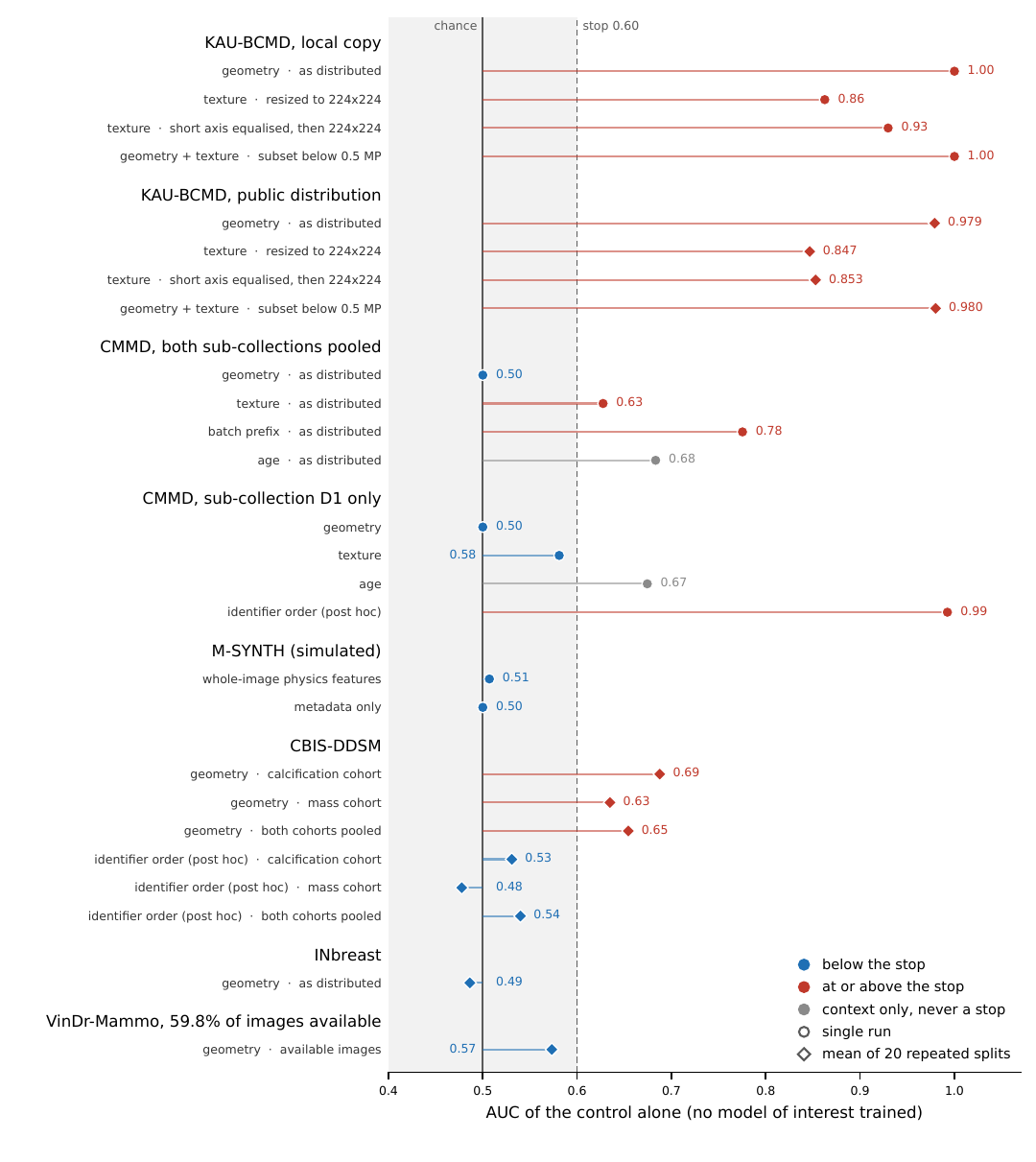}
\caption{\textbf{Pre-training controls on six public mammography datasets.} Each marker is the AUC of one control alone against the diagnostic label (x axis, ``AUC of the control alone (no model of interest trained)''); the dashed line marks the 0.60 stop. Marker colour encodes the verdict: blue, below the stop; red, at or above the stop; grey, context (the single-covariate control, which is reported and never stops a study). Circles are single runs and diamonds are means over 20 repeated splits; the value is printed beside each marker. Blocks from top to bottom: KAU-BCMD local copy (geometry; texture after resizing; texture after short-axis equalisation; geometry + texture on the subset below 0.5 megapixels); KAU-BCMD public Kaggle archive (the same four rows); CMMD pooled (geometry; texture; batch prefix; age); CMMD D1 (geometry; texture; identifier order; age); M-SYNTH (whole-image physics features; metadata only); CBIS-DDSM (geometry: calcification, mass, pooled; identifier order: calcification, mass, pooled); INbreast (geometry); VinDr-Mammo available subset (geometry). Single runs: the KAU-BCMD local-copy rows (one patient-grouped five-fold cross-validation) and the CMMD and M-SYNTH rows (one fit scored on a frozen test split; the batch-prefix value is computed from the label file). Means over 20 repeated patient-grouped five-fold splits: the KAU-BCMD public-archive rows and the CBIS-DDSM, INbreast and VinDr-Mammo rows. The identifier-order rows were added post hoc, for CMMD D1 after the registered controls had passed that cohort, and are exploratory. The CBIS-DDSM verdicts rest on the 20-split mean, a statistic chosen after the results were seen; on the single official splits the mass and calcification cohorts are below the stop. Only the geometry control was run on INbreast and VinDr-Mammo, and geometry and identifier order on CBIS-DDSM, and the VinDr-Mammo row describes 11,965 of the 20,000 released images. Values, SDs and ranges are in Table 4.}
\label{fig:2}
\end{figure}

\begin{table}[!tbp]
\centering\footnotesize
\caption{\textbf{Every control, every dataset and configuration, as measured.} ``Single run'' means one fit at \nolinkurl{random_state = 0}; ``20-split mean'' means the mean out-of-fold AUC over 20 repeated patient-grouped five-fold splits, with SD and range in brackets.}
\label{tab:4}
\begin{tabularx}{\textwidth}{>{\raggedright\arraybackslash}X>{\raggedright\arraybackslash}X>{\raggedright\arraybackslash}X>{\raggedright\arraybackslash}X>{\raggedright\arraybackslash}X}
\toprule
\textbf{Dataset and configuration} & \textbf{Control} & \textbf{AUC} & \textbf{Scoring} & \textbf{Verdict} \\
\midrule
KAU-BCMD, local copy, as held & geometry & 1.0000 & single run (one patient-grouped five-fold cross-validation) & fail \\
KAU-BCMD, local copy, resized to 224 × 224 & texture & 0.8625 & single run (as above) & fail \\
KAU-BCMD, local copy, short axis equalised to 467 px, then 224 × 224 & texture & 0.9297 & single run (as above) & fail \\
KAU-BCMD, local copy, subset below 0.5 MP & geometry + texture & 1.0000 & single run (as above) & fail \\
KAU-BCMD, public Kaggle archive, as distributed & geometry & 0.9790 (SD 0.0082; 0.9712--0.9995) & 20-split mean & fail \\
KAU-BCMD, public Kaggle archive, resized to 224 × 224 & texture & 0.8465 (SD 0.0110; 0.8233--0.8641) & 20-split mean & fail \\
KAU-BCMD, public Kaggle archive, short axis equalised to 467 px, then 224 × 224 & texture & 0.8529 (SD 0.0097; 0.8342--0.8743) & 20-split mean & fail \\
KAU-BCMD, public Kaggle archive, subset below 0.5 MP & geometry + texture & 0.9800 (SD 0.0112; 0.9649--1.0000) & 20-split mean & fail \\
CMMD, pooled D1 + D2 & geometry & 0.5000 & single run, frozen test split & pass \\
CMMD, pooled D1 + D2 & texture & 0.6275 & single run, frozen test split & fail \\
CMMD, pooled D1 + D2 & batch prefix (D1 against D2) alone & 0.7753 & single computation from the label file & fail \\
CMMD, pooled D1 + D2 & age alone & 0.6832 & single run, frozen test split & context, never a stop \\
CMMD D1 only, splits re-frozen & geometry & 0.5000 & single run, frozen test split & pass \\
CMMD D1 only, splits re-frozen & texture & 0.5810 & single run, frozen test split & pass \\
CMMD D1 only, splits re-frozen & identifier order {[}post hoc{]} & 0.9926 & single run, frozen test split (20-split mean 0.9871; Table 8) & fail (control added post hoc) \\
CMMD D1 only, splits re-frozen & age alone & 0.6744 & single run, frozen test split & context, never a stop \\
M-SYNTH, frozen test split, n = 352 & whole-image physics features & 0.5070 & single run, frozen test split & pass \\
M-SYNTH, frozen test split, n = 352 & metadata only (density + dose) & 0.5000 & single run, frozen test split & pass (0.50 by construction) \\
CBIS-DDSM, calcification cohort & geometry & 0.6876 (SD 0.0089; 0.6669--0.7040) & 20-split mean & fail on the mean (statistic chosen post hoc; Table 9) \\
CBIS-DDSM, mass cohort & geometry & 0.6349 (SD 0.0060; 0.6243--0.6438) & 20-split mean & fail on the mean (statistic chosen post hoc; Table 9) \\
CBIS-DDSM, calcification + mass pooled, grouped by patient across cohorts & geometry & 0.6543 (SD 0.0053; 0.6419--0.6643) & 20-split mean & fail on the mean (statistic chosen post hoc; Table 9) \\
CBIS-DDSM, calcification cohort & identifier order {[}post hoc{]} & 0.5308 (SD 0.0139; 0.5059--0.5501) & 20-split mean (official split 0.5299) & pass (control added post hoc) \\
CBIS-DDSM, mass cohort & identifier order {[}post hoc{]} & 0.4777 (SD 0.0154; 0.4444--0.4961) & 20-split mean (official split 0.4451) & pass (control added post hoc) \\
CBIS-DDSM, calcification + mass pooled, grouped by patient across cohorts & identifier order {[}post hoc{]} & 0.5399 (SD 0.0099; 0.5247--0.5570) & 20-split mean & pass (control added post hoc) \\
INbreast, as distributed, BI-RADS 1/2 against 4/5 & geometry & 0.4863 (SD 0.0217; 0.4367--0.5153) & 20-split mean & pass \\
VinDr-Mammo, available subset (11,965 of 20,000 images), BI-RADS 1/2 against 4/5 & geometry & 0.5732 (SD 0.0041; 0.5654--0.5797) & 20-split mean & pass \\
\bottomrule
\end{tabularx}
\end{table}

Three datasets fail at least one stop-role control: KAU-BCMD, CMMD and CBIS-DDSM. Three pass every control that was run on them: M-SYNTH (two controls), INbreast (geometry only) and the available subset of VinDr-Mammo (geometry only). CMMD restricted to D1 passes the two registered stop controls and fails the identifier-order control added post hoc; it is not counted among the passes. CBIS-DDSM passes the identifier-order control in both cohorts and pooled (Section 3.4) and remains counted among the failures on geometry.

\subsection{KAU-BCMD: the class is a function of the file header}

\begin{table}[!tbp]
\centering\footnotesize
\caption{\textbf{Image dimensions by class in the labelled KAU-BCMD folder (local copy).} }
\label{tab:5}
\begin{tabularx}{\textwidth}{>{\raggedright\arraybackslash}Xl>{\raggedright\arraybackslash}X>{\raggedright\arraybackslash}X}
\toprule
\textbf{Class} & \textbf{Images} & \textbf{Distinct image sizes} & \textbf{Dimensions (count)} \\
\midrule
BI-RADS 1 (negative) & 1,865 & 3 & 2816 × 3584 (1,552) · 1284 × 577 (237) · 824 × 578 (76) \\
BI-RADS 4 (positive) & 102 & 2 & 467 × 578 (98) · 482 × 578 (4) \\
BI-RADS 5 (positive) & 24 & 24 & every one of the 24 images has a unique size \\
Shared between classes & \textasciitilde{} & 0 & no positive image has the same dimensions as any negative image \\
\bottomrule
\end{tabularx}
\end{table}

In the local copy every image above 5 megapixels is BI-RADS 1, 1,552 of them, and every BI-RADS 4 image is 0.27 megapixels. The negatives occupy three sizes; the positives are small and, for BI-RADS 5, individually sized. This pattern is what separate export or cropping procedures for the two classes would produce; we did not observe how the files were made. The class is recoverable from the file header, and the geometry control returns AUC 1.0000.

Image content does not explain the separation. Visual inspection at matched display size shows that both classes are whole-breast craniocaudal and mediolateral-oblique views and not lesion crops (Figure 1). What differs is how the files were produced: in the initial analysis the positives carried systematically more high-frequency energy per pixel than the negatives.

\begin{figure}[!tbp]\centering
\includegraphics[width=\textwidth]{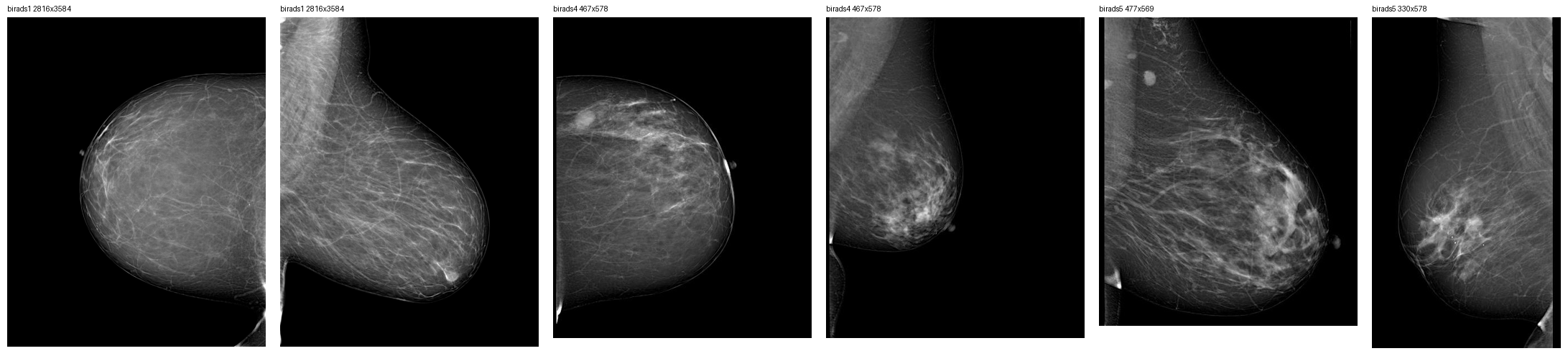}
\caption{\textbf{Sample images by class from the local KAU-BCMD copy.} Two images from each of the three BI-RADS classes of the labelled folder, shown at matched display size with the native pixel dimensions printed above each panel (BI-RADS 1: 2816 × 3584; BI-RADS 4: 467 × 578; BI-RADS 5: 477 × 569 and 330 × 578). Both classes are whole-breast craniocaudal and mediolateral-oblique views and not lesion crops. In this copy the negative class occupies three image sizes (2816 × 3584, n = 1,552; 1284 × 577, n = 237; 824 × 578, n = 76), the BI-RADS 4 images two (467 × 578, n = 98; 482 × 578, n = 4), each of the 24 BI-RADS 5 images has a unique size, and no positive image shares its dimensions with any negative image; a classifier given only width, height, megapixels and aspect ratio separates the classes at AUC 1.0000 under patient-grouped five-fold cross-validation. Images are from the King Abdulaziz University Breast Cancer Mammogram Dataset (KAU-BCMD; Alsolami et al.~\citep{r25}), licensed under CC BY 4.0, and are resized for display.}
\label{fig:1}
\end{figure}

\textbf{The public Kaggle archive shows the same association.} The archive was compared file by file with its extracted copy: all 2,378 images match in folder, file name, dimensions and byte size. Its BI-RADS 1 dimensions and counts are identical to Table 5. Its BI-RADS 4 images are 1267 × 578 (98) and 1282 × 578 (4): the same counts and heights as the local copy but exactly 800 pixels wider. The local folder was therefore a different export of the same images and not a transcription of the archive; the local copy was a PNG folder produced from a DICOM export. The 24 BI-RADS 5 images again have 24 distinct sizes, all 183--477 pixels wide. No dimension is shared between the classes in the archive either.

\begin{table}[!tbp]
\centering\footnotesize
\caption{\textbf{Geometry control on the public KAU-BCMD archive, BI-RADS 1 against 4/5.} All rows are means over 20 repeated patient-grouped five-fold splits.}
\label{tab:6}
\begin{tabularx}{\textwidth}{>{\raggedright\arraybackslash}Xlll>{\raggedright\arraybackslash}Xl}
\toprule
\textbf{Setting} & \textbf{Images} & \textbf{Mean AUC} & \textbf{SD} & \textbf{Range over 20 splits} & \textbf{Verdict} \\
\midrule
All patients & 1,991 & 0.9790 & 0.0082 & 0.9712--0.9995 & fail \\
Excluding the 21 patients with images in both classes & 1,903 & 0.9397 & 0.0320 & 0.8791--0.9980 & fail \\
Grouped by case number only (64 case numbers recur under two year prefixes) & 1,991 & 0.9801 & 0.0051 & 0.9713--0.9882 & fail \\
\bottomrule
\end{tabularx}
\end{table}

The archive value is below the local 1.0000 for an identifiable reason. One patient holds all four 1282-pixel BI-RADS 4 images. A width of 1282 lies between the positive width 1267 and the negative width 1284, so in any split that places this patient in the test fold the classifier has never seen that width and ranks the four images with the negatives. They are misranked in every split, and no other image is; a lookup on exact image size, with no fitting, gives 1.0000. In the local export, where the BI-RADS 4 width is 467 pixels and well below every negative width, a single threshold separates the classes, which is consistent with the 1.0000 measured there.

\textbf{Preprocessing did not remove the association.} On the local copy, resizing everything to 224 × 224 removes the geometry channel and leaves a texture difference that still separates the classes at 0.8625; equalising to a common 467-pixel short axis before resizing gave 0.9297; and the resolution-matched subset returned 1.0000, because no true match exists: the ``matched'' negatives are at 0.48 megapixels against 0.27 megapixels for the positives. On the public archive the three remedies give 0.8465 after resizing (SD 0.0110, range 0.8233--0.8641), 0.8529 after short-axis equalisation (SD 0.0097, range 0.8342--0.8743) and 0.9800 on the subset below 0.5 megapixels (SD 0.0112, range 0.9649--1.0000), which there holds 76 BI-RADS 1 and 24 BI-RADS 5 images from 28 patients. On the local copy, short-axis equalisation scored higher than plain resizing. That observation does not replicate on the public archive, where the two differ by 0.006, within one SD, and it is not claimed here. Our interpretation, which we did not test further, is that resampling cannot give the two classes a common history: it cannot restore information that the negatives do not have at the common scale, nor add to them the compression signature that the positives carry.

\textbf{Scale, had the headers been uninformative.} The local copy holds 501 patients and 1,000 breasts, of which 60 are BI-RADS 4--5 (50 and 10 respectively) against 938 BI-RADS 1. With a 25\% test split that is about 15 positive test breasts, for which the smallest detectable AUC difference is roughly 0.15, far above the 0.02 that the planned benchmark targeted. The local copy could not have supported that benchmark even with uninformative headers. The counts 938 / 50 / 10 exclude the 2 case-side units whose folder labels conflict; forcing a label onto those 2 units gives 939 / 51 / 10. Both sets of counts are kept.

\textbf{Categories absent.} BI-RADS 0 and BI-RADS 2 are absent from the local copy, and its BI-RADS 3 folder holds a single file that does not match the naming convention. The public archive also has no BI-RADS 0 or BI-RADS 2 folder; its BI-RADS 3 folder holds 387 images.

\subsection{CMMD: the sub-collection prefix predicts the diagnosis}

\begin{table}[!tbp]
\centering\footnotesize
\caption{\textbf{CMMD by sub-collection, from the label file.} }
\label{tab:7}
\begin{tabularx}{\textwidth}{llll>{\raggedright\arraybackslash}X>{\raggedright\arraybackslash}X}
\toprule
\textbf{Sub-collection} & \textbf{Benign} & \textbf{Malignant} & \textbf{Total} & \textbf{Malignant fraction} & \textbf{Molecular subtype recorded (fraction)} \\
\midrule
D1 & 544 & 563 & 1,107 & 0.508582 & 0.000000 \\
D2 & 12 & 753 & 765 & 0.984314 & 0.979085 \\
\bottomrule
\end{tabularx}
\end{table}

D1 is 50.9\% malignant and D2 is 98.4\% malignant; molecular subtype is recorded for 97.9\% of D2 and for none of D1. The prefix alone predicts malignancy at AUC 0.7753 (0.775303 at full precision, re-derived from the label file). The two-part structure is described by the dataset's authors \citep{r26}. What we measure is its consequence: any analysis that pools D1 and D2 without controlling for the sub-collection inherits a channel worth AUC 0.7753.

The geometry control returns exactly 0.5000 on CMMD, and the manifest shows why: all 5,202 images have identical pixel dimensions of 2294 × 1914, so the geometry feature vector is constant and the control is at chance by construction. Bit depth is not uniform: 5,200 images are stored at 8 bits and 2 at 16. An earlier check on a sample of two series had recorded the archive as uniform at 8 bits; both observations are kept.

The texture control failed on the pooled dataset at 0.6275 against the 0.60 threshold, and that failure is what triggered the investigation of the sub-collections. After restriction to D1 and re-freezing of the splits, geometry is 0.5000 and texture is 0.5810, both below the stop. We read the D1 texture value as a floor and not as a defect, on the untested assumption that global texture statistics covary with breast density; under that reading a model scoring at or below 0.58 on D1 has shown nothing beyond what four global statistics provide.

Age alone scores 0.6832 on the pooled dataset and 0.6744 on D1. Both values are reported as context and neither stops a study: CMMD is a diagnostic cohort in which age is associated with malignancy, which is a property of the population. The value is nonetheless useful to a reader of any CMMD result, because it is the performance a model must exceed before it has shown that it uses the image.

\subsection{CMMD D1: identifier order predicts the diagnosis (post hoc control)}

\begin{table}[!tbp]
\centering\footnotesize
\caption{\textbf{Identifier-order control on CMMD D1.} The only feature is the serial number in the patient identifier. This control was added post hoc and is exploratory.}
\label{tab:8}
\begin{tabularx}{\textwidth}{>{\raggedright\arraybackslash}Xl>{\raggedright\arraybackslash}X}
\toprule
\textbf{Evaluation} & \textbf{Breasts} & \textbf{AUC} \\
\midrule
Frozen D1 validation split & 171 & 0.9700 \\
Frozen D1 test split & 331 & 0.9926 \\
20 repeated patient-grouped five-fold splits & 1,107 & mean 0.9871 (SD 0.0009; range 0.9852--0.9892) \\
\bottomrule
\end{tabularx}
\end{table}

The malignant fraction by decile of serial number, from the lowest serial numbers to the highest, is 0.54, 0.36, 0.89, 0.43, 0.00, 0.00, 0.00, 0.88, 1.00 and 0.98. The released label file is thus ordered in long runs of one diagnosis, and a classifier that sees nothing but a record's position recovers the label almost perfectly.

Three consequences follow. First, CMMD restricted to D1 passes the registered geometry and texture controls (0.5000 and 0.5810) and fails this control; it is not a clean negative control. Second, the result shows inside this paper that a pass on the registered controls is not a certificate. Third, the released gate now specifies a fifth control, identifier order, and states that it was added post hoc.

An identifier is not a property of the image, and a model given pixels alone cannot read it directly. Whether any image property covaries with serial order, for example through acquisition period, was not tested. The same leakage was found in a separate benchmark by the authors, where it led to the withdrawal of an arm that used CMMD D1.

The same control on CBIS-DDSM (also post hoc) stays below the stop: mean AUC 0.5308 for the calcification cohort (1,511 images, 753 patients), 0.4777 for the mass cohort (1,592 images, 892 patients) and 0.5399 pooled (3,103 images, 1,566 patients), with every one of the 60 split values at or below 0.5570; on the official splits it gives 0.5299 (calcification, 284 test images) and 0.4451 (mass, 361 test images). The malignant fraction by serial decile is uneven in the calcification cohort (0.34 to 0.46 in deciles 1--9, 0.66 in decile 10), but the identifier alone does not separate the classes. The CBIS-DDSM label files therefore do not carry the CMMD ordering defect, and the CBIS-DDSM verdict rests on geometry alone.

\subsection{CBIS-DDSM: a geometry--label association above the stop}

Every full-field image was located from its own DICOM header (10,229 DICOM objects read; 2,853 full mammograms identified by series description and patient identifier) and joined to the official case-description files. The label is image-level pathology: malignant if any annotated lesion on the image is malignant, otherwise benign. The official per-cohort splits were kept, and patients were grouped across the two cohorts.

\begin{table}[!tbp]
\centering\footnotesize
\caption{\textbf{Geometry control on CBIS-DDSM.} The mean over 20 repeated splits is the verdict statistic and was chosen after the results were seen; the official single-split value is given beside it.}
\label{tab:9}
\begin{tabularx}{\textwidth}{l>{\raggedright\arraybackslash}X>{\raggedright\arraybackslash}X>{\raggedright\arraybackslash}X>{\raggedright\arraybackslash}X}
\toprule
\textbf{Cohort} & \textbf{Images / patients} & \textbf{Mean AUC over 20 splits (range)} & \textbf{Verdict on the mean} & \textbf{Official split only} \\
\midrule
Calcification & 1,261 / 635 & 0.6876 (0.6669--0.7040) & fail & 0.5625 (37 test images); below the stop \\
Mass & 1,592 / 892 & 0.6349 (0.6243--0.6438) & fail & 0.5884; below the stop \\
Pooled & 2,853 / 1,460 & 0.6543 (0.6419--0.6643) & fail & 0.6182 (31 overlapping patients removed); above the stop \\
\bottomrule
\end{tabularx}
\end{table}

The association is weaker than in KAU-BCMD and is not a lookup: the classes share 122 image dimensions in the pooled cohort, and 2,448 distinct sizes occur among 2,853 images. It is above the stop in all 60 repeated splits. The verdict nevertheless depends on a choice made after the results were seen. On the single official splits the mass cohort scores 0.5884 and the calcification cohort 0.5625, the latter on only 37 test images, and both are below the stop; the pooled cohort still exceeds it at 0.6182 with the 31 overlapping patients removed. We therefore describe the result as a failure of the geometry control under the repeated-split statistic, and not as a defect of the dataset. The mechanism was not investigated in this study, and nothing here attributes the association to a particular acquisition, digitisation or curation step.

One feature of the copy bounds the calcification result. The calcification test split is almost absent from it, 37 of 284 official breast-views, so the calcification row describes mainly the official training images.

The two official cohorts also interact with the official splits. Once the calcification and mass cohorts are combined, 31 patients appear in both the official training and the official test split, 18 in calcification-training and mass-test and 13 the other way round, and 29 breast-views are in both splits; within either cohort alone there is no overlap. Any analysis that pools the two cohorts while keeping the official splits places those patients on both sides. Patient-level leakage on CBIS-DDSM arising from evaluation protocols has been quantified by others \citep{r15}.

\subsection{Datasets that passed the controls run on them}

M-SYNTH passes both available controls: whole-image physics features score 0.5070 and metadata alone, breast density and dose label, scores 0.5000, the latter 0.50 by construction.

INbreast passes the geometry control at a mean of 0.4863 (range 0.4367--0.5153). Its 379 analysed images take only two sizes, 2560 × 3328 and 3328 × 4084, and both occur in both classes.

The available VinDr-Mammo subset passes at 0.5732 (range 0.5654--0.5797), the pass closest to the stop; 21 image dimensions are shared between the classes. This value describes 11,965 of the 20,000 released images, with the available share between 57.9\% and 69.7\% in every cell of split by BI-RADS category. It is not a property of the complete release, and the subset may be selected in ways the control cannot see.

Only the geometry control was run on INbreast and VinDr-Mammo, and only two controls exist for M-SYNTH. A pass on one or two controls is not a clean bill of health, as Section 3.4 shows for a dataset that passed two.

\subsection{Registered items that were not completed}

The second publisher route for KAU-BCMD was not checked, the archive was not obtained under an institutional account, and it is not established which export produced the local folder or whether the 800-pixel width difference reflects cropping, padding or a different release.

The RSNA challenge dataset \citep{r38} and BreastMNIST \citep{r39} were not gated. The texture, batch/site, covariate and bit-depth controls were not run on CBIS-DDSM, INbreast or VinDr-Mammo.

The registration stated that the generalisation claim would be withdrawn if all five further datasets passed. Under the registered single-split design the CBIS-DDSM mass and calcification cohorts are each below the stop, and only the pooled cohort, a configuration we constructed, exceeds it. Whether that condition is met therefore depends on the post hoc statistic or on counting the pooled configuration. We claim only that the controls returned values above the stop on some configurations of a third dataset.

The initial controls were not re-run over repeated splits. The two texture values at 0.6275 and 0.5810 lie on either side of the 0.60 stop, and dispersion for them could change a verdict; the batch prefix at 0.7753 does not share that weakness.

The downstream cost of a failing control, meaning the change in a conventional classifier's AUC with and without the leaking channel, was not measured.

\begin{center}\rule{0.5\linewidth}{0.5pt}\end{center}

\section{Discussion}

\subsection{What the controls cost and what they are}

The geometry control reads four integers per file and decodes no pixels. For a dataset of about 2,000 images, such as the 1,991-image KAU-BCMD inventories, the classifier fit takes seconds. Reading the headers is the larger cost when files sit in cloud storage: for a 10,000-object dataset it takes minutes (the notebooks ran in about four minutes for the 10,229 CBIS-DDSM objects and about 25 minutes for the VinDr-Mammo subset). The texture control decodes pixels and is slower. On the three datasets with a failure the controls returned 1.0000 on the local KAU-BCMD copy and 0.979 on the public archive, 0.654 on pooled CBIS-DDSM under the repeated-split statistic and, through the batch control, 0.7753 on CMMD. Set against the cost of a benchmark built on a compromised cohort, the cost of the controls is small. That is the argument for running them at intake and not only when something looks wrong.

Two properties of the design are separable from its implementation. The controls are impoverished by construction: they receive inputs that should not carry diagnostic content, so a high score is hard to interpret as anything but leakage. This makes them more informative than a performance ceiling or a learning curve, either of which can be explained by model capacity. And the covariate control is reported and not enforced, because a gate that stopped a study on the ground that age predicts malignancy in a diagnostic cohort would be measuring epidemiology.

\subsection{Resizing is not a defence, and the mechanisms differ}

On KAU-BCMD, resizing to a standard input size did not neutralise the association. It removed the geometry channel and left a texture difference that separates the classes at 0.8625 on the local copy and 0.8465 on the public archive; equalising the short axis first did not help on either copy (0.9297 locally, 0.8529 on the public archive). In this dataset the confound is carried by the information content of the image and not only by its dimensions, so matching dimensions addresses what a reader can see and leaves the cause in place.

CMMD shows a different mechanism with the same consequence. The archive is dimensionally uniform and returns chance on geometry. What fails is that two sub-collections with 50.9\% and 98.4\% malignancy are distributed together, so that the identifier prefix alone reaches AUC 0.7753. Neither the geometry control nor the batch control would have caught the other's dataset, which is the argument for more than one control. CBIS-DDSM shows a third pattern: no single dimension separates the classes, yet the joint distribution of size and shape is associated with the label above the stop under repeated splits. A check that only looked for image sizes unique to one class, without scoring a classifier, would not have flagged it.

\subsection{A pass is not a certificate}

CMMD D1 is the instructive case. It was adopted as the corrected cohort after the pooled dataset failed, it passed both registered stop controls, and it remained the working cohort until the identifier-order control was added. A control that had not been anticipated, the serial number in the identifier, then returned a test AUC of 0.9926. A set of controls can only detect the channels its authors anticipated; a single untuned learner gives a lower bound on leakage and not an upper one. The appropriate reading of every ``pass'' in Table 4 is that no leakage was detected by the controls that were run, and for INbreast and VinDr-Mammo that means one control.

The identifier-order control was added after the fact, and adding controls after a result is known is itself a forking path. We report it as exploratory for that reason, and we do not claim that five controls are sufficient where four were not.

\subsection{Implications for users of these datasets}

The KAU-BCMD association is present in the public Kaggle archive identified in Table 3, so it bears on any study trained on that archive, and a reader can run the check on their own copy. We do not extend the statement to other distributions of the dataset. The CMMD findings are properties of the released label file: the sub-collection prefix and the serial number are in the identifier of every row, and an analysis that pools D1 and D2, or that uses record order within D1 in any way, inherits the corresponding channel. For CBIS-DDSM, geometry is associated with pathology above the stop under repeated splits but not on the official mass and calcification splits, and pooling the two official cohorts under the official splits puts 31 patients on both sides.

The practical recommendation is narrow and testable: report the control AUCs in any paper that trains on a public mammography dataset; state the sub-collection composition when pooling; check patient overlap whenever official splits from different cohorts are combined; and treat record order and identifiers as potential label carriers.

\subsection{Context within a quantum-versus-classical programme}

This paper belongs beside a set of quantum-versus-classical comparisons only because the controls were written to protect them. Its relation to the proposed framework for quantum advantage \citep{r36} is limited to one observation: where the framework asks that the correctness of an output be validated, a dataset whose label is readable from its file header prevents that validation for every arm, and the deficiency precedes anything the computation does. The framework is a proposal by authors affiliated with quantum-computing vendors, and we adopt none of its forecasts. This paper reports no quantum advantage, no speed-up and no quantum-enhanced performance, offers no quantum result as evidence of any benefit, and there is no quantum component in the inference path.

\begin{center}\rule{0.5\linewidth}{0.5pt}\end{center}

\section{Limitations}

\begin{enumerate}
\def\labelenumi{\arabic{enumi}.}
\tightlist
\item
  \textbf{The KAU-BCMD archive is one of two registered routes, and its provenance is incomplete.} The association is confirmed in one Kaggle-hosted archive, identified here by file name, size and member dates. The hosting page, uploader and version were not recorded at download. The archive's file name matches the identifier of the Kaggle page given in the dataset paper's data-availability statement \citep{r25} (https://www.kaggle.com/asmaasaad/king-abdulaziz-university-mammogram-dataset), which is consistent with the dataset authors' own deposit but does not establish it; a re-upload by a third party is not excluded. The second registered route was not checked.
\item
  \textbf{The KAU-BCMD analysis of the local copy was not registered.} It was an ad hoc diagnostic, the gate was formalised afterwards, and the result is exploratory.
\item
  \textbf{The registration is internal.} It is a dated written protocol with a deviations log, released with the code, and not a record in a public registry.
\item
  \textbf{The batch/site and identifier-order controls are post hoc additions.} The CMMD sub-collection confound was found by investigation after the texture control failed, and the identifier-order control was added after the registered gate had passed CMMD D1, then applied to CBIS-DDSM. The five-control gate released here is therefore partly retrospective in design.
\item
  \textbf{The initial results are single runs at one seed.} No dispersion is available for them, and the two texture values at 0.6275 and 0.5810 lie on either side of the stop.
\item
  \textbf{The verdict statistic for the follow-up results was chosen after the results were seen.} The follow-up results carry 20-split dispersion and their verdict is taken on the mean. On the single official splits the CBIS-DDSM mass cohort scores 0.5884 and the calcification cohort 0.5625 (37 test images), both below the stop, while the pooled cohort exceeds it at 0.6182 with the 31 overlapping patients removed.
\item
  \textbf{For CBIS-DDSM the premise of the geometry control is weaker.} In a digitised-film dataset cropped to the breast, image size may partly reflect breast size or the digitiser or site, neither of which was tested, so for CBIS-DDSM the premise that geometry carries no diagnostic information is weaker than for the fixed-format digital datasets.
\item
  \textbf{A single untuned classifier gives a lower bound on leakage.} A control that fails has failed; a control that passes means only that this learner, on these features, found nothing. The gate is a one-sided screen and a pass is not a certificate, as the CMMD D1 result shows.
\item
  \textbf{Protocols are not uniform across datasets.} The local KAU-BCMD copy is cross-validated once, CMMD and M-SYNTH are scored once on a frozen split, and the follow-up analyses are scored over 20 repeated cross-validation splits. The available controls differ because the recorded metadata differ, and only the geometry control was run on CBIS-DDSM, INbreast and VinDr-Mammo.
\item
  \textbf{The scripts that produced the initial KAU-BCMD and CMMD D1 values are not available.} The released gate script is a reconstruction from their written specification and is not the code that produced those values. The raw inputs and the measured outputs are released; the transformation between them is documented in prose for those rows. The follow-up results are produced end to end by the released notebooks, saved with their outputs.
\item
  \textbf{The registered checksums of the pooled CMMD splits cannot be verified.} The pooled split files were replaced when the protocol was amended to D1 and are not recoverable. Both the registered and the realised checksums are reported in the release.
\item
  \textbf{Six datasets are a small base for a general recommendation.} KAU-BCMD and CMMD are the two datasets in which the first failures were found, so that part of the sample is not independent of the finding. CBIS-DDSM, INbreast and VinDr-Mammo were named in the registration before they were examined.
\item
  \textbf{Three sizes are on record for KAU-BCMD and they disagree.} The dataset paper reports 1,416 cases and 5,662 images \citep{r25}; the IEEE DataPort record of the dataset lists 1,521 cases and 6,109 images; and a benchmark compilation that includes the dataset \citep{r33} lists 1,416 cases and 2,206 JPG images. We confirmed the first against the abstract of the dataset paper and the third against an excerpt of the compilation's text. The second is as transcribed during our initial dataset verification and was not re-checked for this manuscript. The disagreement is unresolved. It affects no control AUC, but the denominator of any statement about the dataset as a whole is uncertain.
\item
  \textbf{CMMD bit depth is not uniform.} A full pass over the manifest finds 5,200 images at 8 bits and 2 at 16, against an earlier sampled check that recorded 8 bits throughout. Both values are kept, and a bit-depth check is added to the gate.
\item
  \textbf{BI-RADS is a radiologist-assigned category and not pathology.} The KAU-BCMD, INbreast and VinDr-Mammo labels are a different kind of object from the pathology labels of CMMD and CBIS-DDSM, and BI-RADS 0 is a workflow action and not a tissue state.
\item
  \textbf{The follow-up copies are incomplete or partly characterised.} VinDr-Mammo is 59.8\% of the release; the CBIS-DDSM copy lacks 247 of 284 official calcification-test breast-views; the INbreast access agreement is not on file. Each result describes the copy analysed.
\item
  \textbf{Patient identity in KAU-BCMD is inferred from file names.} 64 case numbers recur under two year prefixes; grouping by case number alone changes the archive AUC by 0.001 (Table 6), but the patient structure is not documented by the publisher.
\item
  \textbf{The texture definitions were fixed for the follow-up re-run.} The initial analysis did not define them numerically, so the local and public texture AUCs are not demonstrably computed in the same way.
\item
  \textbf{Not assessed.} Whether the controls detect confounds other than geometry, texture, batch, identifier order and a single covariate; any dataset outside mammography; and the downstream performance cost of a failing control.
\item
  \textbf{Threshold choice.} The verdicts depend on the threshold only near it. Re-scoring the 24 stop-role results at 0.55, 0.65, 0.70 and 0.75 changes the number of failing results from 14 at 0.60 to 16, 12, 10 and 10. The failures of both KAU-BCMD copies and of pooled CMMD do not depend on the threshold anywhere in that range, because each rests on at least one value of 0.7753 or above. Three dataset-level verdicts do depend on it: CBIS-DDSM fails on geometry for any threshold up to 0.6876 and passes at 0.70; VinDr-Mammo, which passes at 0.60 with a mean of 0.5732, would fail at 0.55; and CMMD D1, which passes its registered controls at 0.60, would fail them at 0.55 through its texture value of 0.5810. Every value is reported in full, so a reader who prefers another threshold can re-score each dataset directly.
\end{enumerate}

\begin{center}\rule{0.5\linewidth}{0.5pt}\end{center}

\section{Conclusion}

File-level controls run before training returned values far above chance on copies of KAU-BCMD and on pooled CMMD, and above a 0.60 stop on CBIS-DDSM under a statistic chosen post hoc. A cohort that passed the registered controls then failed one added later. The controls are inexpensive, they detect only what they were built to look for, and their results, including the unremarkable ones, are worth reporting alongside any model trained on these datasets.

\begin{center}\rule{0.5\linewidth}{0.5pt}\end{center}

\section*{Data availability}
The derived tables that support this article are deposited in a public Zenodo repository, doi:10.5281/zenodo.24372683. The deposit contains no mammographic images. Figure 1 reproduces six images from KAU-BCMD under that dataset's CC BY 4.0 licence.

The deposit comprises: \nolinkurl{gate_summary.csv} (every control result with its source); \nolinkurl{threshold_sensitivity.csv} and \nolinkurl{threshold_sensitivity_summary.json} (every verdict re-scored at thresholds from 0.55 to 0.75, and the near-chance split spread); \nolinkurl{gate_results_d1.csv} and \nolinkurl{gates_msynth.csv} (CMMD D1 and M-SYNTH control results); \nolinkurl{gate_results_d1_identifier_order.csv} and \nolinkurl{gate_results_d1_identifier_order_splits.csv} (the post hoc identifier-order control on CMMD D1, frozen-split values and the 20 per-split values); \nolinkurl{gate_results_cbis_identifier_order.csv} and \nolinkurl{gate_results_cbis_identifier_order_splits.csv} (the same control on CBIS-DDSM, per cohort and pooled, with the 60 per-split values); \nolinkurl{cmmd_labels.csv}, \nolinkurl{cmmd_manifest.csv} and \nolinkurl{cmmd_batch_table.csv} (CMMD label table, per-image manifest and sub-collection table); \nolinkurl{split_train.csv}, \nolinkurl{split_val.csv} and \nolinkurl{split_test.csv} (frozen CMMD D1 splits); \nolinkurl{kau_bcmd_labelled_inventory.csv}, \nolinkurl{kau_bcmd_local_inventory_redacted.csv} and \nolinkurl{kau_geometry_table.csv} (local KAU-BCMD inventories and dimension table; export-folder names in the local inventory are replaced by opaque labels); \nolinkurl{kau_bucket_inventory.csv} (per-image width, height, bit depth, compression, file size and SHA-256 for all 2,378 images of the public KAU-BCMD archive); \nolinkurl{cbis_split_overlap_patients.csv} (the 31 CBIS-DDSM patients in both official splits); \nolinkurl{checksums.sha256} and \nolinkurl{verification_log.md} (checksums and the re-derivation record); and \nolinkurl{01_preregistration.md} (the internal protocol with its dated deviations log). Per-image inventories exist for the public KAU-BCMD copy only; none was produced for CBIS-DDSM, INbreast or VinDr-Mammo.

The source datasets are obtained by the reader from their providers. KAU-BCMD \citep{r25} was obtained as the Kaggle-hosted archive \nolinkurl{king-abdulaziz-university-mammogram-dataset.zip}, recorded as CC BY 4.0. The file name matches the Kaggle page given in the dataset paper \citep{r25} (https://www.kaggle.com/asmaasaad/king-abdulaziz-university-mammogram-dataset); the version and the download date were not recorded. CMMD \citep{r26,r27} is distributed by The Cancer Imaging Archive \citep{r28} under CC BY 4.0 (https://doi.org/10.7937/tcia.eqde-4b16). CBIS-DDSM \citep{r29} is distributed by The Cancer Imaging Archive under CC BY 3.0 and the archive's data-usage policy. VinDr-Mammo \citep{r31} is distributed by PhysioNet under the PhysioNet Restricted Health Data License 1.5.0, with credentialed access. INbreast \citep{r30} is obtained from its providers on request under an access agreement; that agreement is not on file with the authors at the time of writing. M-SYNTH \citep{r32} is distributed by the U.S. Food and Drug Administration as Regulatory Science Tool RST24AI06.01 under CC0.

\section*{Code availability}
The code is deposited in the same repository (doi:10.5281/zenodo.24372683). It comprises \nolinkurl{acceptance_gates.py} (the gate in reusable form); \nolinkurl{patchfeat.py} (the original M-SYNTH control code); \nolinkurl{verify_numbers.py} (re-derivation of the quoted counts from the tables); \nolinkurl{threshold_sensitivity.py} (re-scoring of every verdict at other thresholds); \nolinkurl{cmmd_d1_identifier_order.py} and \nolinkurl{cbis_ddsm_identifier_order.py} (the post hoc identifier-order control on CMMD D1 and on CBIS-DDSM; the latter reads the four public CBIS-DDSM case-description files, which the reader obtains from the archive); \nolinkurl{make_fig2_gate_panel.py} (Figure 2); five notebooks saved with their outputs, \nolinkurl{kau_bcmd_comparison.ipynb}, \nolinkurl{kau_bcmd_preprocessing_controls.ipynb}, \nolinkurl{cbis_ddsm_geometry.ipynb}, \nolinkurl{inbreast_geometry.ipynb} and \nolinkurl{vindr_mammo_geometry.ipynb}; \nolinkurl{run_all.sh}; and \nolinkurl{requirements.txt}, \nolinkurl{requirements_lock.txt} and \nolinkurl{ENVIRONMENT_MANIFEST.md} (software versions and machine).

\nolinkurl{acceptance_gates.py} implements the procedure as specified in writing and is not the code that produced the initial KAU-BCMD and CMMD D1 values; the scripts that produced those values are not available. The identifier-order control (\nolinkurl{cmmd_d1_identifier_order.py}) uses the same classifier on the released split files and reproduces the recorded age value on the same split as a check of protocol.

\section*{Author contributions}
A.T.: investigation, software, data curation, validation, writing -- review and editing. R.S.: investigation, validation, writing -- review and editing. M.Y.E.H.: conceptualisation, methodology, formal analysis, supervision, writing -- original draft, writing -- review and editing.

\section*{Ethics statement}
This study used only publicly available, de-identified or simulated datasets (KAU-BCMD, CMMD, CBIS-DDSM, INbreast, VinDr-Mammo and M-SYNTH). The authors performed no human-subject research, collected no patient data and required no ethics approval.

\section*{Declaration of AI use}
LLM-based assistants were used under the authors' supervision to edit and restructure author-written text, to check references and quoted numbers, and to prepare the submission files. All content was reviewed and approved by the authors, who take full responsibility for the manuscript.

\impact{The controls described here are meant to stop models from being trained and reported on datasets whose labels can be recovered from non-diagnostic information, which protects patients, clinicians and regulators from performance claims that say nothing about disease. Two misuses are possible. A failed control could be read as a statement that a dataset or its creators are at fault, whereas the findings describe properties of particular released copies and label files, several of them chosen or added post hoc; and a pass could be read as a certificate of quality, which the paper explicitly rejects. No images or new patient data are released; the deposit contains code and derived tables only.}

\acks{The authors thank the creators and maintainers of the public datasets analysed here. This work was supported by Qanary Inc.~internal funds. \textbf{Competing interests.} All authors are employees of Qanary Inc., a company developing mammography screening software. M.Y.E.H. and A.T. are named inventors on a pending US patent application (no. 19/876,140, filed 17 November 2026) concerning input-integrity gating. The authors declare no other competing interests.}

\end{document}